\documentclass[conference]{IEEEtran}
\IEEEoverridecommandlockouts
\usepackage{cite}
\usepackage{amsmath,amssymb,amsfonts}
\usepackage{algorithmic}
\usepackage{graphicx}
\usepackage{textcomp}
\usepackage{xcolor}
\usepackage{lipsum}
\usepackage{multicol}
\usepackage{subcaption}
\usepackage{soul}
\usepackage{bm}
\usepackage{float}
\def\BibTeX{{\rm B\kern-.05em{\sc i\kern-.025em b}\kern-.08em
    T\kern-.1667em\lower.7ex\hbox{E}\kern-.125emX}}
\begin{document}
\title{Bayesian Inference on Binary Spiking  Networks Leveraging Nanoscale Device Stochasticity}
\author{\IEEEauthorblockN{Prabodh Katti, Nicolas Skatchkovsky, Osvaldo Simeone, Bipin Rajendran, Bashir M. Al-Hashimi}
\IEEEauthorblockA{{Department of Engineering},
King's College London, London, UK \\
Email: bipin.rajendran@kcl.ac.uk}
}

\maketitle

\begin{abstract}
Bayesian Neural Networks (BNNs) can overcome the problem of overconfidence that plagues traditional frequentist deep neural networks, and are hence considered to be  a key enabler for reliable AI systems. However, conventional hardware realizations of BNNs are resource intensive, requiring the implementation of random number generators for synaptic sampling. Owing to their inherent stochasticity during programming and read operations,  nanoscale memristive devices can  be directly leveraged for sampling, without the need for additional hardware resources.  In this paper, we introduce a novel Phase Change Memory (PCM)-based hardware implementation for BNNs with  binary synapses. The proposed architecture consists of separate weight and noise planes, in which PCM cells are configured and operated to represent the nominal values of weights and to generate the required noise for sampling, respectively. Using experimentally observed PCM noise characteristics, for the exemplary Breast Cancer Dataset classification problem, we obtain hardware accuracy and expected calibration error matching that of an 8-bit fixed-point (FxP8) implementation, with projected savings of over 9$\times$ in terms of core area transistor count.

\end{abstract}


\section{Introduction}
 Modern neural networks tend to produce overconfident decisions, misrepresenting the inherent epistemic uncertainty that arises from access to limited data \cite{guo2017calibration}. In contrast, critical applications need a well-calibrated measure of confidence in the prediction for risk-aware decision-making \cite{shukla2020mc,blundell2015weight}. Bayesian neural networks (BNNs) offer a principled solution to this problem by encoding epistemic uncertainty in a probability distribution on the model parameter space  and by implementing ensemble predictors \cite{wilson2020case,jang2021bisnn}. Implementing inference via BNNs require  a source of stochasticity, i.e., a random number generator, in order to sample from the probability distribution over the model parameters. This paper investigates the  idea of leveraging nanoscale device-level programming noise as a source for sampling in Bayesian inference.

\noindent \emph{Related Work:} The proposed approach  contrasts with most existing BNN implementations, which rely on pseudo-random number generators (PRNGs), yielding resource-intensive systems \cite{yang2020all}. For example, CMOS-based implementations proposed in \cite{cai2018vibnn,shukla2020mc}  require static random access memory (SRAM) cells, multipliers and analog-to-digital converters (ADCs),  entailing  large area footprints. Furthermore, they  rely on  conventional von Neumann architecture with physically separate memory and processing units.
\begin{figure}
\centering
\begin{subfigure}{0.47\textwidth}
 \centering   \includegraphics[width=.5\textwidth]{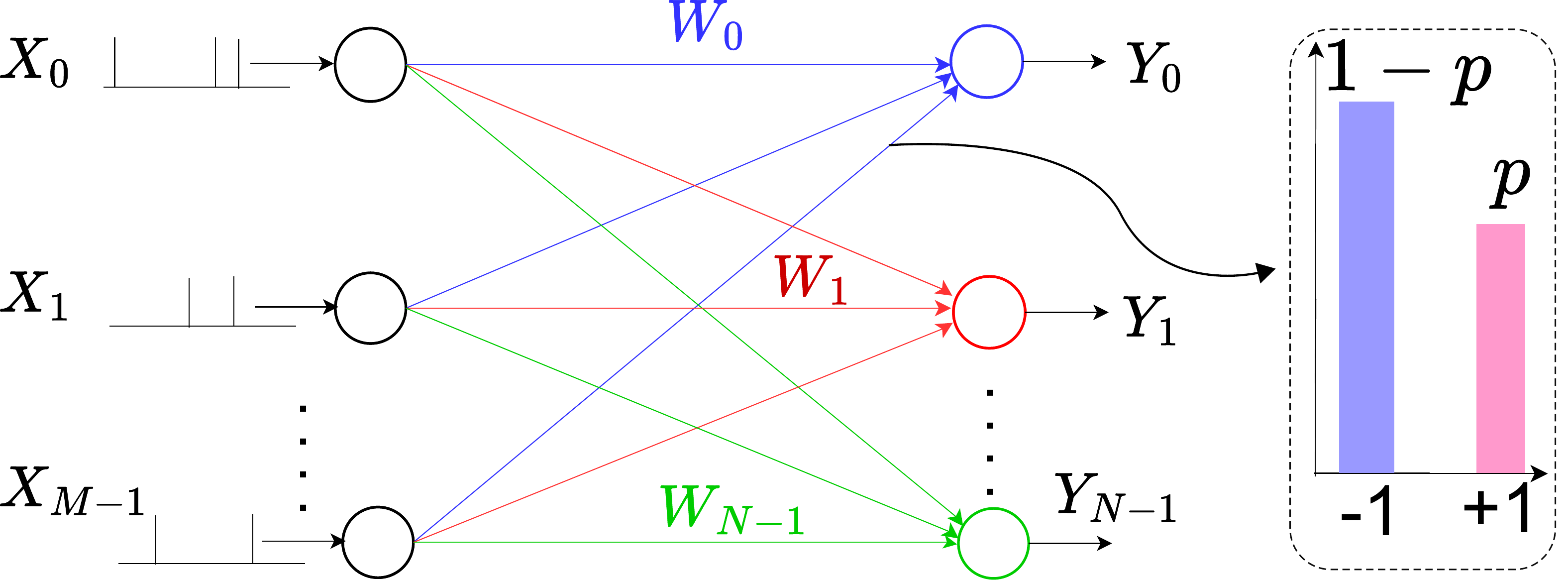}
    \caption{Bayesian spiking neural network with binary weights obtained during inference by sampling from a Bernoulli  distribution. } 
    \label{fig:subnetwork}
\end{subfigure}

\begin{subfigure}{0.47\textwidth}
   \centering \includegraphics[width=\textwidth]{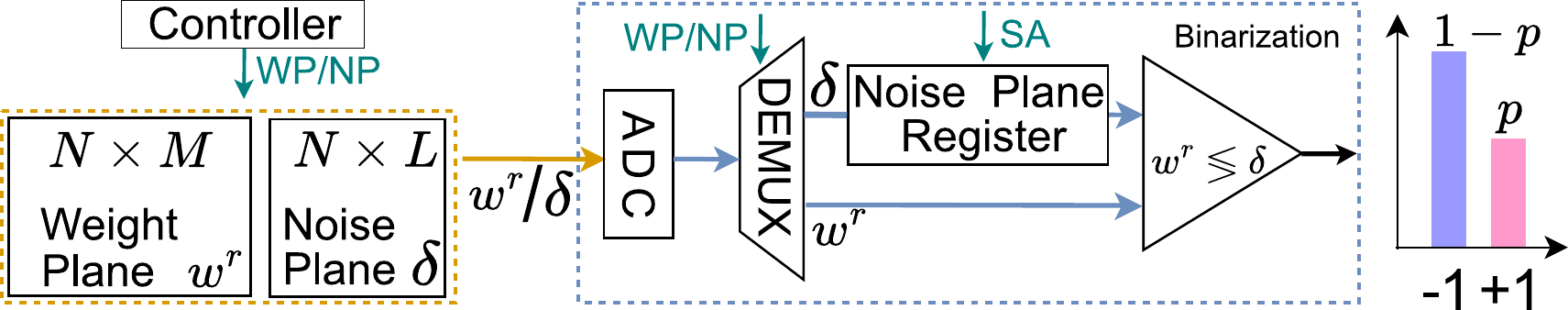}
    \caption{Proposed hardware architecture consisting of a $N\times M$ cross-bar of differential PCM (DPCM) cells, referred to as the weight plane (WP), which stores the nominal weights $w^r$, and of a noise plane (NP) with $N \times L$ DPCM cells, which generates randomness for synaptic sampling. We choose $L<M$ and reuse the conductance values from the $L$ rows in the noise plane stored in a register through stochastic arbitration (SA), in order to reduce sampling correlation while optimizing area. }

    \label{fig:block}
\end{subfigure}

\caption {Illustration of a Bayesian binary SNN (BSNN) and its implementation using PCM crossbar.} 
\label{hierarchy2}
\end{figure}

Non-volatile memory (NVM) devices such as Resistive RAM (RRAM), Phase Change Memory (PCM) and Spin-Transfer Torque RAM (STTRAM)  are being explored for the implementation of in-memory computing (IMC) architectures to circumvent the von Neumann bottleneck \cite{li2019overview}.  
These nanoscale devices support information storage by modulating their effective electrical conductance in smaller footprints, enabling multi-level and even analog computing \cite{joshi2020accurate,doi:10.1063/1.5042408}. However, it is known that these devices are characterised by significant programming and read noise levels \cite{joshi2020accurate}. 

Implementations of BNNs via IMC include \cite{dalgaty2021situ}, which uses an RRAM array with training implemented via Markov Chain Monte Carlo (MCMC), which is time and resource intensive \cite{salimans2015markov}. 
The authors of \cite{yang2020all} propose to use the Bernoulli distributed noise of STT-RAM   to generate Gaussian random variables. This approach generally requires a large number of devices in order to ensure that the sum of their contributions can be averaged to approach a Gaussian density, as dictated by the central limit theorem.

Spiking Neural Networks (SNNs) \cite{jang2019introduction} have been widely investigated for their energy efficiency, with hardware implementations ranging from digital to mixed-signal IMC platforms including industry-fabricated TrueNorth and Loihi chips \cite{davies2021advancing,rajendran2019low,truenorth}. Bayesian learning algorithms for SNNs were proposed  in \cite{skatchkovsky2022bayesian} (see also references therein), although their hardware implementation has not been addressed so far.

\noindent The main contribution of this paper are as follows:\\
\noindent $\bullet$   We introduce a novel hardware implementation of binary Bayesian SNNs (BSNNs, see Fig. 1(a)) by means of an IMC architecture that utilizes the inherent programming stochasticity of PCM devices for synaptic sampling. The proposed architecture and design, illustrated in Fig. 1(b), allow software-trained network weights to be directly transferred to hardware, while minimizing the amount of additional hardware resources introduced for sampling.

\noindent $\bullet$ Using experimentally observed PCM noise characteristics, for the exemplary Breast Cancer Dataset classification task, we obtain hardware accuracy and expected calibration error matching that of an 8-bit fixed-point implementation, with over 9$\times$ projected savings in terms of core area transistor count.

\section{Background}\label{sec: Background}
In this section, we review preliminary material on PCM devices, as well as on binary BSNNs.

\noindent \emph{PCM Devices:} A phase change memory (PCM) device consists of a chalcogenide material such as Ge$_2$Sb$_2$Te$_5$ sandwiched between two metal electrodes. 
The device conductance can be programmed to a desired state within a pre-determined target bound by applying a combination of electric pulses, varying the number,  amplitude, and width of the programming pulse which modifies the proportion and configuration of crystalline and amorphous phase of the material between the two electrodes\cite{doi:10.1063/1.5042408,rajendran2019building}, allowing analog IMC\cite{joshi2020accurate,papandreou2011programming}.
Such devices are then configured as a differential pair to allow storing of  positive and  negative weight values within a crossbar array  to represent a synaptic network. 

PCM devices exhibit noise during programming and read operations. Referring to Fig. 2, we note the following essential features of the inherent  device noise obtained from  measurements of over 10,000 devices fabricated in 90 nm CMOS  in \cite{joshi2020accurate}: (\emph{i}) The programming   noise $\sigma_p(G)$ is approximately Gaussian distributed with a standard  deviation  depending quadratically on the mean conductance $G$; (\emph{ii}) The read noise noise $\sigma_r(G)$ is time-dependent, and it is of smaller magnitude as compared to the programming noise. \vspace{.05in}

\begin{figure}
\centering
\begin{subfigure}{0.19\textwidth}
    \includegraphics[width=\textwidth]{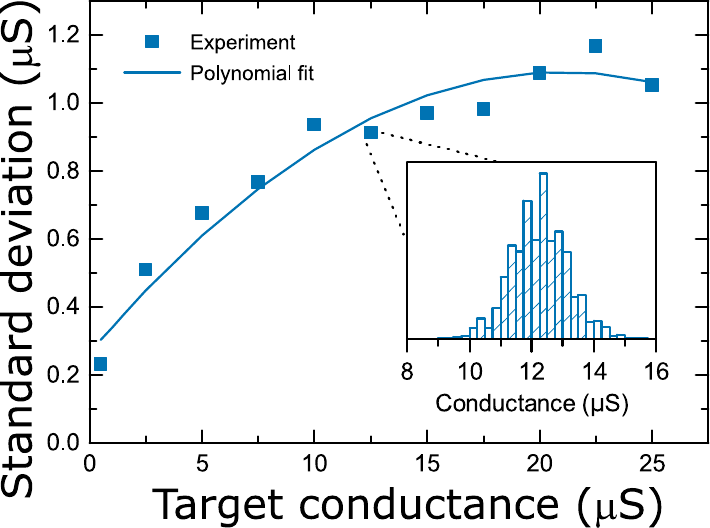}
    \caption{Programming noise}
    \label{fig:wrnoit}
\end{subfigure}
\hfill
\begin{subfigure}{0.24\textwidth}
    \includegraphics[width=.84\textwidth]{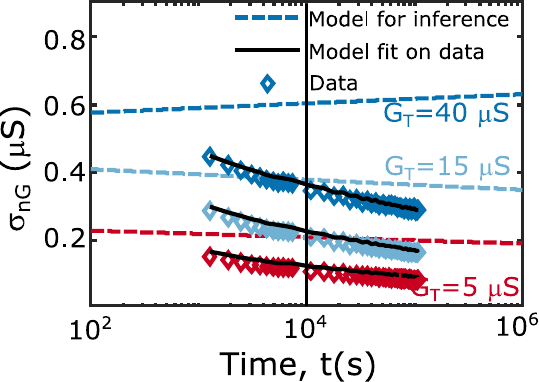}
    \caption{Read noise}
    \label{fig:rdnoit}
\end{subfigure}
\label{hierarchy}
\caption{Noise characteristics of PCM, adapted from \cite{joshi2020accurate}. } 
\label{fig:noises}
\end{figure}


\noindent \emph{Bayesian SNNs with Binary-Valued Synapses:}
In this work, we consider BSNNs with binary-valued synapses. Their operation is characterized by synaptic weights $w$ that can assume only the states $\{-1,+1\}$. Accordingly, the synaptic weights are described  by a vector $w \in \{-1,+1\}^{|w|}$, with $|w|$ denoting the size of the vector. During training, a mean-field distribution $q(w)$, describing the distribution of independent Bernoulli random variables, is optimized via an online generalization of the Bayesian learning rule \cite{khan2021bayesian} introduced in \cite{jang2021bisnn}. 

The result of this optimization is a real-valued vector $w^{\text{r}} \in \mathbb{R}^{|w|}$, which determines the distribution $q(w)$. Specifically, the random vector $w\sim q(w)$ consists of independent random variables, with each $i$th entry $w_i$ being equal to $+1$ with probability $p_i=\textrm{sigmoid}(2 w^r_i)$, or equivalently $w_i^r = \frac{1}{2}\log\big( p_i/(1-p_i))$, with $p_i$ and $w^r_i$ being the $i$th entries of vectors $p$ and $w$. 

Given vector $w^r$, a sample $w\sim q(w)$ can be approximately generated  via the Gumbel-Softmax trick\cite{jang2016categorical}
\begin{equation}
    \label{eqn: soft-binar}
    w = \tanh\bigg(\frac{w^r+\delta}{\tau}\bigg),
\end{equation}
where the function is applied element-wise, and each entry of vector $\delta$ is generated independently as $\frac{1}{2} \log\big({\epsilon}/{(1-\epsilon)}\big)$, with $\epsilon \sim\mathcal{U}(0,1)$ being a Gumbel-distributed random variable. In the limit $\tau \rightarrow 0$,  the sample $w$ follows the distribution $q(w)$. 

During inference, several samples $w\sim q(w)$ are generated in order to obtain an ensemble of predictors, whose decisions are averaged to produce the final decision. The neuronal dynamics is implemented using leaky integrate and fire (LIF)  with discrete spike response model (see, e.g., \cite{kaiser2020synaptic}).

\section{PCM-Based Hardware Architecture}\label{sec:architecture}
In this section, we introduce the proposed PCM-based hardware architecture that implements an ensemble predictor via BSNNs with binary synapses.

\noindent\emph{Overview of the architecture}:\label{sec:architecture-overview} As illustrated in Fig. 1(b) and Fig. 3, the proposed architecture consists of two planes: the weight plane, implemented by an $N \times M$ cross-bar of differential PCM (DPCM)  devices, and a noise plane, given by an $N \times L$ cross-bar of DPCM devices. The weight plane implements the nominal weights $w^r$, while the noise plane is used to generate the random vector $\delta$ for the sampling step (\ref{eqn: soft-binar}). The key idea is to implement the noise plane by leveraging the programming noise of PCM devices.  To this end, the main challenge is to engineer the  noise plane so as to approximate the sampling step (\ref{eqn: soft-binar}) by using the characteristics of PCM devices reviewed in the previous section.

Provisioning the noise plane to have the same dimensions as the weight plane, i.e., setting $L=M$, would allow the noise plane to produce independent random weights as dictated by  (\ref{eqn: soft-binar}). To improve the overall area efficiency, a smaller noise plane with $L<M$ is proposed, by reusing the noise plane values multiple times (chosen via a stochastic arbitration scheme) for binarization. This, however, results in correlated weights, causing an approximation error as compared to (\ref{eqn: soft-binar}). The proposed architecture supports a flexible choice of the parameter $L$, and we study the effect of $L$ on performance and efficiency via simulations.

To ensure that the same sample weights are applied throughout the presentation of all the spikes for an input signal, we sample and hold the conductance values of the devices in a noise plane register bank (Fig. 1b). Moreover, this allows the implementation of a stochastic arbitration scheme that randomly chooses one column of $N$ conductances among the $L$ columns stored this register bank for binarization of the weights from the weight plane, thereby reducing the effect of correlation    (see Sec. IV).

\noindent\emph{Leveraging PCM noise}:  As demonstrated in \cite{joshi2020accurate}, PCM programming noise  is Gaussian distributed, with state-dependent mean and standard deviations (see Fig. 2(a)). We propose to leverage this randomness to generate the desired   vector $\delta$ for the sampling step \eqref{eqn: soft-binar}. To this end, we approximate the Gumbel distribution with a zero-mean Gaussian distribution with standard deviation  $\sigma_\delta \approx 0.8$. A sample from this distribution  is obtained by appropriately scaling and combining the conductance of PCM devices, as discussed next. 
  
\noindent\emph{Weight plane and noise   plane}:
To motivate our design choice of separating weight and noise plane, consider first a simpler solution in which a single DPCM device, with conductance levels $G^+$ and $G^-$, is used to represent the desired Gaussian distribution with mean equal to the nominal weight $w^r$ and variance $\sigma^2_\delta$. With some abuse of notation, we take $w^r$ here to represent any element of vector $w^r$. To this end, the two PCM devices must satisfy the constraints
\begin{align}
    w^{r}& = G^{+} - G^{-}, \label{eq:mean}\\
    \sigma_{\delta}^2& = \sigma_{p}^2 (G^{+})+ \sigma_{p}^2 (G^{-}) + \sigma_{r}^2 (G^{+}) + \sigma_{r}^2 (G^{-}), \label{eq:sd} 
\end{align}
 where the programming- and read-noise standard deviations $\sigma_{p}$ and $\sigma_{r}$ are as described in the previous section (see Fig. \ref{fig:noises}), while the conductance levels $G^+$ and $G^-$ lie in the $0$-$25$ $\mu$S  dynamic range of the PCM device. We have empirically observed that this feasible space does not permit the satisfaction of the conditions \eqref{eq:mean} and \eqref{eq:sd} across a useful range of values for the weight $w^r$.

To address this problem, we propose implementing the  the nominal value $w^r$ of a weight  using a differential pair $(G_{w^r}^{+},G_{w^r}^{-})$ within the weight plane; while the sampling noise is obtained by arbitrarily selecting one  cell from a collection of $L$ DPCM cells $(G_{ni}^{+},G_{ni}^{-})$, with $i=1,2,\ldots L$, in the noise plane. We refer to the next subsection for details on the realization on the PCM core. The weight-plane noise plane devices are configured to satisfy the conditions\begin{align}
    w^{r} &= G_{ni}^{+} - G_{ni}^{-} + \kappa^{-1}\big(G_{w^r}^{+} - G_{w^r}^{-}\big),  \label{eq:mean1}\\
        \sigma_{\delta}^2 &= \sigma_{p}^2 (G_{ni}^{+}) + \sigma_{p}^2 (G_{ni}^{-})+\sigma_{r}^2 (G_{ni}^{+}) + \sigma_{r}^2 (G_{ni}^{-})+\nonumber\\
    &\qquad  \kappa^{-2}\bigg( \sigma_{p}^2 (G_{w^r}^{+})+ \sigma_{p}^2 (G_{w^r}^{-}) + \sigma_{r}^2 (G_{w^r}^{+}) + \sigma_{r}^2 (G_{w^r}^{-})\bigg)\nonumber\\
   & \approx \sigma_{p}^2 (G_{ni}^{+}) + \sigma_{p}^2 (G_{ni}^{-})+\sigma_{r}^2 (G_{ni}^{+}) + \sigma_{r}^2 (G_{ni}^{-}),\label{eq:mean2}
 \end{align}where $\kappa>1$ is a suitable scaling parameter, which should be sufficiently large to justify the approximation \eqref{eq:mean2}.  By programming  $ G_{ni}^{+}   \approx G_{ni}^{-}$, we ensure that the conductances of the noise-plane devices do not affect the overall nominal value for a weight, contributing only to the sampling noise. 

As an implementation detail, nominal weights that satisfy the inequality $|w^r|>2$ are  set to a fixed high value, with minimal impact on binary sampling via (\ref{eqn: soft-binar}). Furthermore, based on the available range of device conductances, we   choose  $\kappa=8$ in (4)-(5), and the maximum value of the nominal weights are mapped to the weight plane conductances with a value of $16\,\mu$S, and ceiled weights are mapped to $24\,\mu$S.

\begin{figure}
  \centering
    \includegraphics[width=0.5\textwidth]{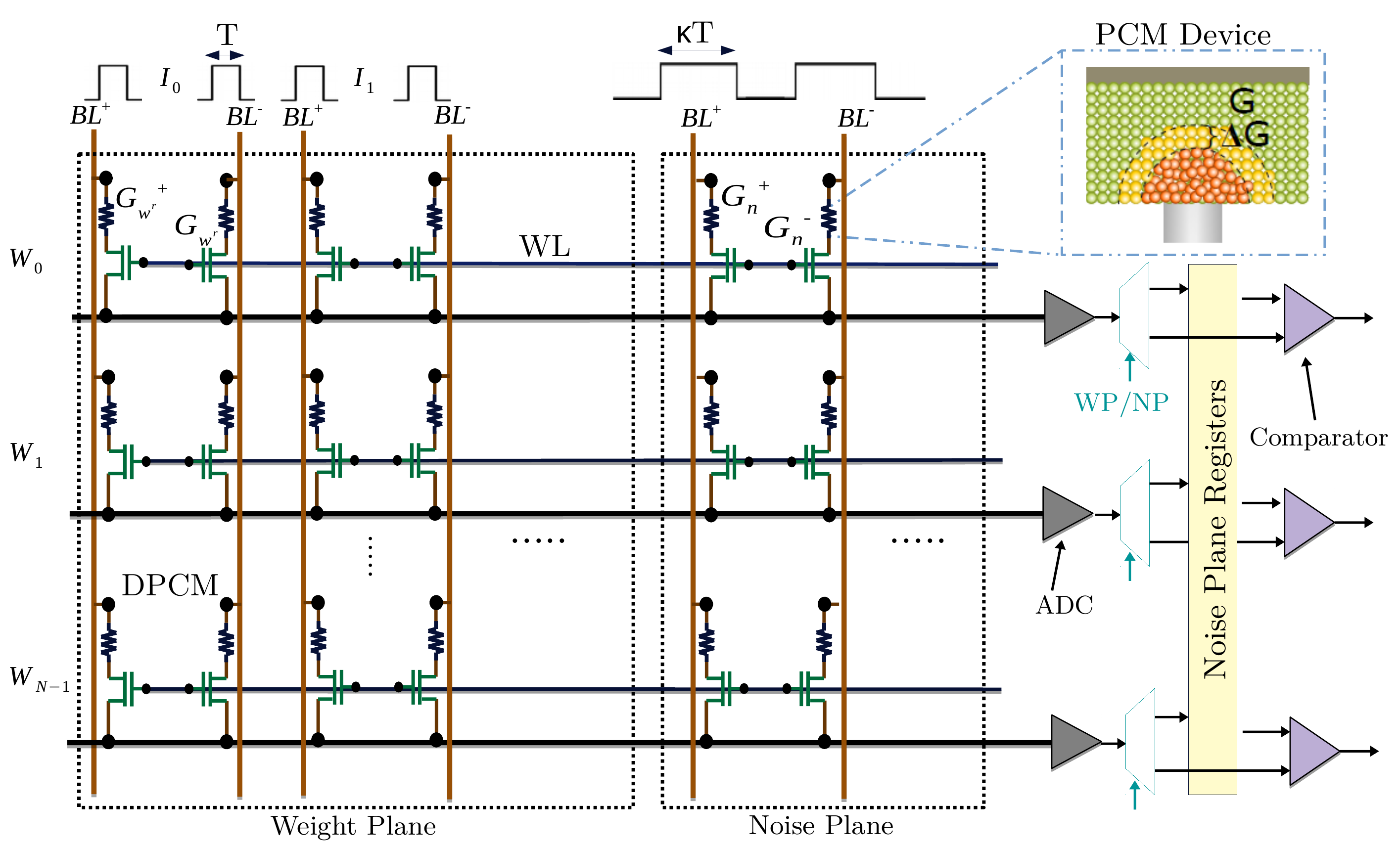}
    \caption{Proposed synaptic core architecture, in which DPCM cells are configured to represent nominal weight values ($w^r$) in the weight plane, as well as to  generate the required noise variables ($\delta$) for Bayesian sampling in the noise plane.}
  \label{fig:crossbar}
\end{figure}

\noindent\emph{Realization on a PCM core}: 
As illustrated in Fig. 3, in the proposed crossbar architecture, the devices for the noise plane are provisioned by allocating  $L$ additional columns. To implement  \eqref{eqn: soft-binar}, we replace  $\tanh(\cdot)$ function with the approximate, hardware-efficient $\text{sign}(\cdot)$ function. {For every batch of inputs, the analog conductances of all the $L$ columns of the DPCM cells from the noise plane are first read and then are discretized and stored in the noise plane register. Since  the product of weights and incoming spikes need to be binarized by using the $\text{sign}(\cdot)$ function before  accumulation in the output neuron, the columns of weight plane are activated one-by-one, and then compared to one of the columns of noise values held in the noise plane register, selected via stochastic arbitration.}

To implement the $\kappa=8$ scaling factor between the nominal weight value and the programmed conductance in the weight plane, we use pulse width modulation along the bit lines  by scaling the duration of the read pulse for the noise plane by a factor of $\kappa$, as compared to the read duration of the weight plane devices (see Fig. \ref{fig:crossbar}). 

To implement forward propagation in the cross-bar with $N$ columns, we require $N$ cycles of the read clock. The access transistors in weight plane are first turned on by enabling the word lines (WL), and read pulses of appropriate width are applied to the bit lines (BL) depending on the presence or absence of a spike at the pre-neuron. We assume an 8-bit ADC for reading the conductance values in both planes. 

The confidence scores are obtained by averaging the rates of the output spiking signals over time, as well as across the ensemble of predictors drawn from the distribution $q(w)$, as detailed in  \cite{jang2021bisnn}. The classification decision is taken by applying a threshold on the confidence scores, and accuracy is determined by comparing the decisions with true labels.

\section{Experiments and conclusions} \label{sec:results}

In this section, we evaluate the proposed architecture by first training the networks in  software via the methodology described in \cite{jang2021bisnn}. Software-trained weights  are then mapped onto a custom-built hardware emulator of a PCM crossbar that supports iterative programming of PCM \cite{joshi2020accurate} by following the approach described in the previous section. 

\begin{figure}[!h]
\centering    \vspace{-.1in}
        \includegraphics[width=0.44\textwidth]{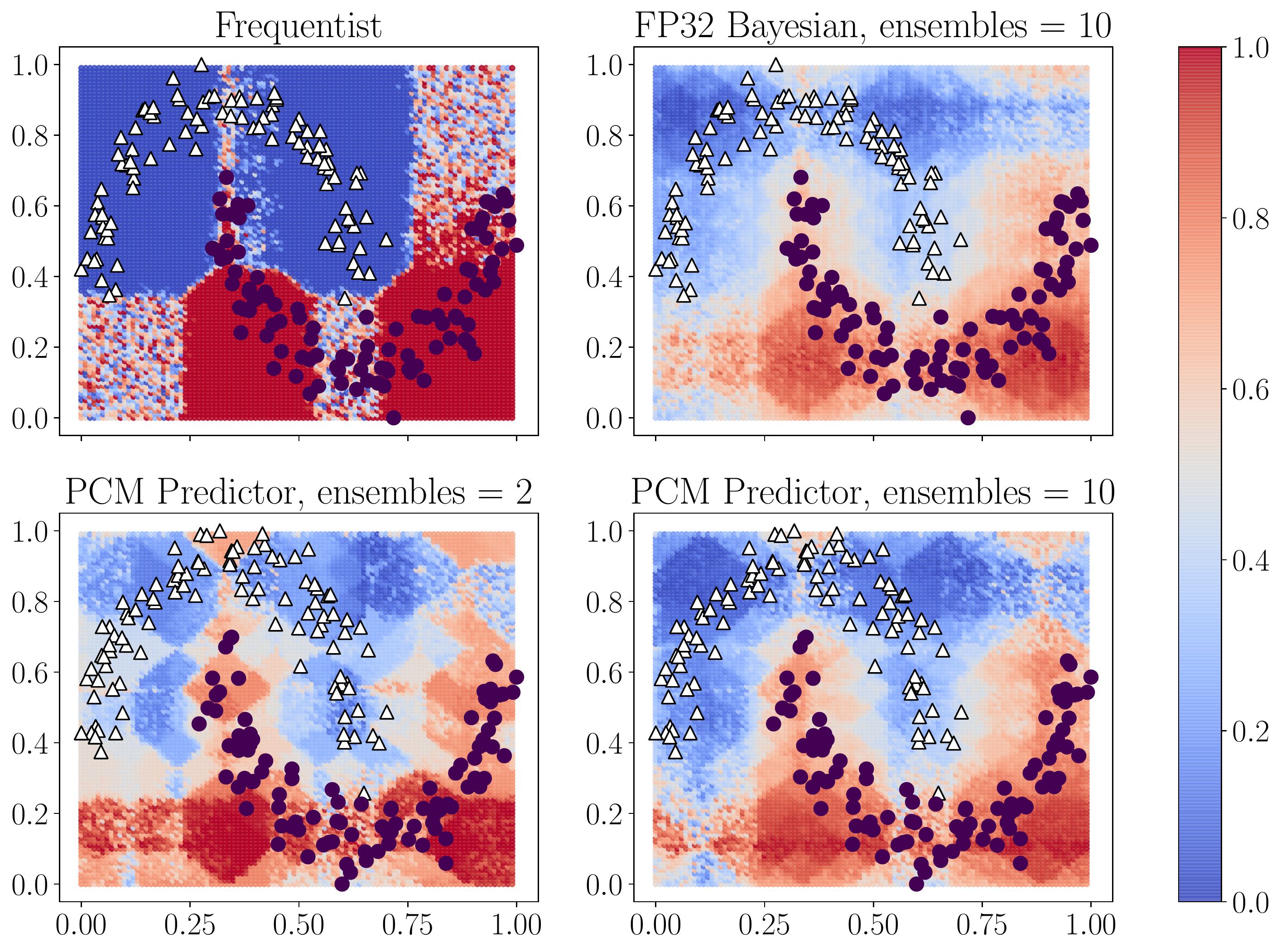}
       \vspace{-.1in}
        \caption{Classification probability  from the PCM-based Bayesian architecture ($L=1$) on the two-moons data set, matches the FP32 Bayesian software baseline.}
        \label{two moons result}
\end{figure}   \vspace{-.1in}
We first consider the two-moons classification problem for which we adopt a  fully connected network with two hidden layers  as in \cite{jang2021bisnn}. In Fig.~\ref{two moons result}, we compare the  confidence scores assigned to the ``circle'' class by the FP32  implementation (top-right) and by the PCM-based implementation (bottom) with different numbers of models, drawn using (\ref{eqn: soft-binar}), in the ensemble. The intensity of the color in these plots refer to the predicted probability that a point belongs to the ``circle'' class. The top two panels show that the Bayesian model is better calibrated than the frequentist implementation (optimized as in \cite{jang2021bisnn}). In the frequentist case, most of the out-of-distribution samples are incorrectly assigned a strong probability of belonging to one of the classes with high confidence, neglecting epistemic uncertainty. The FP32 Bayesian implementation offers  a more reliable quantification of uncertainty by assigning graded probabilities to those points, as can be seen from the hue of colors. Crucially, we note that the PCM implementation, even with $L=1$, (bottom panels) provides a close match to the FP32 baseline with 10 ensembles, with reasonable performance also achieved with just two ensembles.

\begin{figure}[!h]
        \includegraphics[width=0.47\textwidth]{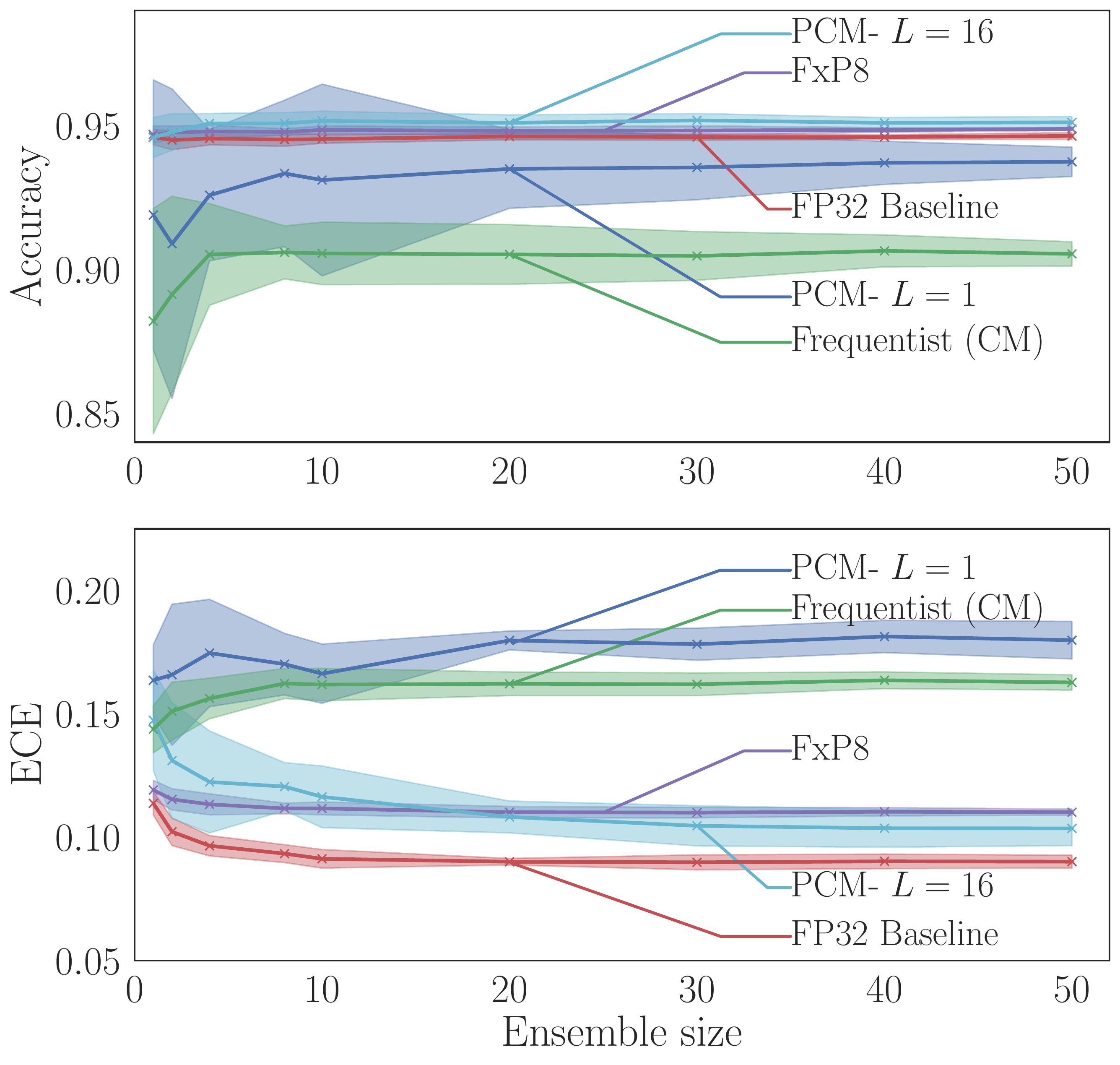}
    
    \caption{Accuracy and expected calibration error (ECE) obtained from the PCM-based Bayesian architecture for the Wisconsin Breast Cancer data set, matching the results from a 8-bit fixed point simulation and outperforming a frequentist (committee machine) implementation.}
    \label{WBCD_result}
\end{figure}

We turn now to a more complex real-world dataset, namely the Wisconsin Breast Cancer data set (Diagnostic), for which we adopt the same network architecture as in \cite{jang2021bisnn}, except that the number of input neurons is 30. The inputs are population coded as in \cite{jang2021bisnn}, and we added a pre-processing step to increase the sparsity of the input data, with an average spiking rate of 0.04 spikes per neuron per time step. Accuracy and the expected calibration error (ECE), defined as in \cite{guo2017calibration}, are plotted as a function of ensemble size in Fig. \ref{WBCD_result}. ECE measures the average discrepancy between the confidence level assigned by the model and the ground-truth.   For the benchmark frequentist implementation, we have adopted a Committee Machine (CM) architecture, whereby the outputs of multiple networks --  all with the same nominal weights trained using standard frequentist learning --  are averaged to eliminate effects of device stochasticity \cite{joksas2020committee}. The ECE and accuracy metrics are shown for different choices of $L$, which determines the size of the noise plane. The main observation from the figure is that setting $L=16$ matches the 8-bit fixed point (FxP8) software benchmark in terms of both accuracy and calibration error. In contrast, with $L=1$, there is a significant performance loss in ECE caused by noise correlation (see Sec. \ref{sec:architecture-overview}).

Based on this observation, we finally estimate the area efficiency of the PCM-based architecture in terms of the transistor count for the synaptic core as compared to a conventional SRAM-based FxP8 implementation. We assume a weight plane with  $256 \times 256$ weights and a noise plane with $256\times 16$ weights. For the CMOS implementation, we assume that a linear-feedback shift register (LFSR)-based PRNG with period long enough is adopted to produce pseudo-random samples. The transistor count in terms of D-Flip Flop is obtained using the approach in \cite{laate2017extended}. To this end, we consider two implementations,  the first in which one 36-bit PRNG serves the entire noise plane, and another in which there is a 22-bit PRNG in each row, in a manner similar to the proposed PCM design. In former case, we estimate a 9.3$\times$ area reduction for the PCM based architecture compared to the CMOS fixed point design, whereas in latter we estimate an 11$\times$ reduction in transistor count.

Future work may explore hardware-aware training to make the network performance robust to non-ideal effects such as the limited endurance and yield of nanoscale memristive devices. 

\label{sec:conc} 

\bibliography{biblio}
\end{document}